\documentclass[runningheads]{llncs}
\usepackage{booktabs}
\usepackage{graphicx}
\usepackage{xcolor}
\usepackage{algpseudocode}
\usepackage{algorithm}
\begin{document}
\setcounter{secnumdepth}{5}
\title {Revolutionizing Disease Diagnosis: A Microservices-Based Architecture for Privacy-Preserving and Efficient IoT Data Analytics Using Federated Learning}

\author{Safa Ben Atitallah\inst{1} \and
Maha Driss\inst{1,2} \and
Henda Ben Ghezala \inst{1}}
\authorrunning{S. Ben Atitallah et al.}
%
\institute{RIADI Laboratory, University of Manouba, Manouba, Tunisia \and
Security Engineering Lab, CCIS, Prince Sultan University, Riyadh, Saudi Arabia}
\maketitle              
\begin{abstract}

Deep learning-based disease diagnosis applications are essential for accurate diagnosis at various disease stages. However, using personal data exposes traditional centralized learning systems to privacy concerns. On the other hand, by positioning processing resources closer to the device and enabling more effective data analyses, a distributed computing paradigm has the potential to revolutionize disease diagnosis. Scalable architectures for data analytics are also crucial in healthcare, where data analytics results must have low latency and high dependability and reliability.
This study proposes a microservices-based approach for IoT data analytics systems to satisfy privacy and performance requirements by arranging entities into fine-grained, loosely connected, and reusable collections. Our approach relies on federated learning, which can increase disease diagnosis accuracy while protecting data privacy. Additionally, we employ transfer learning to obtain more efficient models. Using more than 5800 chest X-ray images for pneumonia detection from a publically available dataset, we ran experiments to assess the effectiveness of our approach. Our experiments reveal that our approach performs better in identifying pneumonia than other cutting-edge technologies, demonstrating our approach's promising potential detection performance.
\end{abstract}

The rapid emergence and evolution of the Internet of Things (IoT) have facilitated the advancement of Machine Learning (ML) and Deep Learning (DL) techniques for healthcare data analytics \cite{qayyum2020secure}. Recent DL and image processing breakthroughs have enabled the development of state-of-the-art computer-aided diagnosis systems, which assist pathologists in obtaining highly accurate and reliable diagnosis results \cite{atitallah2020leveraging}. These cutting-edge systems harness the power of DL and image processing to augment the diagnostic capabilities of healthcare professionals, leading to improved patient care and outcomes \cite{ben2022randomly}.

In this context, we introduce a novel approach based on microservices and Federated Learning  (FL), which can potentially enhance disease detection accuracy while significantly safeguarding data privacy. 
Microservices are designed to be autonomous and decoupled from each other, allowing for flexibility in terms of development, deployment, and updates \cite{atitallah2022microservices,driss2022req}. In addition, they are highly responsive to user demands as they can be designed to be lightweight and optimized for specific tasks \cite{hasan2021sublmume,driss2020servicing}. This means that the system can quickly respond to user requests, providing real-time or near-real-time results, which is crucial in healthcare data analytics, where timely insights are often required. Moreover, microservices can be deployed across multiple edge computing nodes, allowing for distributed processing and reducing the need to transfer large amounts of data to a central location. This can result in lower latency, reduced network overhead, and improved overall system performance.

With FL, collaborative model training is conducted across multiple distributed devices or institutions without sharing raw data  \cite{xu2021federated}. Instead, only model updates are exchanged, ensuring the confidentiality of sensitive information \cite{driss2022federated}. 
The proposed approach harnesses the collective intelligence of multiple experts, resulting in a robust and accurate disease detection model. FL effectively addresses privacy concerns by keeping data localized and minimizing data transfer, ensuring patient data remains secure and protected. 

Besides, Transfer Learning (TL) plays an important role in the healthcare domain to improve the effectiveness and efficiency of various tasks \cite{malik2020comparison}. TL leverages pre-trained models trained on large-scale datasets, such as ImageNet, and applies them to healthcare-specific tasks. 

By leveraging the pre-trained model's knowledge through TL and allowing local fine-tuning on specific target domains, Federated Transfer Learning (FTL) aims to improve the learning performance and generalization capability of models trained in a federated setting \cite{chen2020fedhealth}. It allows for the transfer of learning representations, which can capture relevant features and patterns while preserving local data's privacy.


Our proposed approach embodies a remarkable stride in disease detection, harnessing 
the exceptional scalability and swift response time of microservices, the invaluable privacy-preserving advantages of FL, and the augmentation of model performance, generalization, and prediction accuracy through TL.

The paper outlines several key contributions, which can be summarized as follows:
\begin{itemize}
\item The proposal of a novel microservices-based approach for disease diagnosis.
\item The improvement of privacy in disease diagnosis through applying FL.
\item The utilization of TL to improve model performance in disease diagnosis case studies.
\item The application and assessment of the proposed approach in a real healthcare case study.
\end{itemize}

The structure of the paper is organized as follows. Section 2 provides an overview of the relevant literature and related works in the field of disease detection. Section 3 presents the details of the proposed microservices-based approach for disease detection, including the use of FL and TL. Section 4 covers the implementation details of the proposed approach and presents the experimental analysis conducted to evaluate its performance. Section 5 concludes the study and outlines potential future research directions in disease diagnosis.

\section{Related Works}
This study's literature review emphasizes the significant improvements in healthcare data analytics with the advent and expansion of IoT, ML, and DL techniques. This study mainly deals with the emergence of computer-aided diagnostics systems, which leverage the power of DL and image processing to improve healthcare practitioners' diagnostic capabilities.

\cite{chhikara2020deep} suggested altering the InceptionV3 model, adding five more layers, and using TL to create a DL-based automated diagnostic tool for detecting pediatric pneumonia. With the help of gamma correction and compression, the suggested model's accuracy reached 90.1\%. The suggested model was tested on a standard Women and Children's Medical Center X-ray dataset. Its precision, recall, accuracy, and ROC accuracy scores were compared to those of ResNet, ImageNet, Xception, and Inception. Unlike our approach, which takes advantage of FL  and TL, 
the authors concentrated on preprocessing the dataset and training/testing multiple CNN architectures.

In \cite{kundu2021pneumonia}, a computer-aided diagnosis method for automatically identifying pneumonia using images of chest X-rays was developed. Deep TL was used to deal with the restricted amount of data provided, and an ensemble of three convolutional neural network models—GoogLeNet, ResNet18, and DenseNet121—was generated. The base learner weights were determined using the weighted average ensemble methodology. A five-fold cross-validation approach was used to evaluate the suggested method using two pneumonia X-rays available to the general audience, the Kermany and RSNA datasets. The validation demonstrated that the method outperformed commonly used ensemble approaches and produced results superior to state-of-the-art techniques. This study focused on centralized training and testing of DL models and did not address security and privacy concerns related to handling patients' sensitive data.

A novel approach based on evidence-based fusion theory was suggested in \cite{ben2022fusion}, allowing a series of DL classifiers to be fused for more accurate illness detection findings. This study's essential contribution was using the Dempster-Shafer theory to merge five pre-trained convolutional neural networks for detecting pneumonia from chest X-ray images, including VGG16, Xception, InceptionV3, ResNet50, and DenseNet201. To test this method, experiments were carried out utilizing a publically available dataset, including over 5800 chest X-ray pictures. Compared to existing state-of-the-art approaches, the suggested methodology demonstrated good detection performance. Contrary to our approach, this study focused on a conventional methodology of applying and testing DL models.

The study presented in \cite{islam2022deep} aimed to provide a DL-based framework for detecting pneumonia from X-ray images using Compressed Sensing (CS) data, which could be useful for healthcare professionals working in distant areas. The system was bandwidth-preserving and energy-efficient due to the use of CS, which allowed for effective far-end pneumonia diagnosis. Through rigorous simulations, the suggested technique achieved a pneumonia detection accuracy of 96.48\% even when only 30\% of the samples were communicated. These results demonstrated the effectiveness of the suggested strategy in detecting pneumonia, even with limited information. This study addressed the challenge of applying DL models to a restricted amount of data; it did not address important issues related to the proposed approach's privacy, scalability, and reusability.

A DL architecture based on EfficientNetB7 was proposed in \cite{moussaid2023implementation} to classify medical X-ray and Computed Tomography (CT) images of lungs into three categories: common pneumonia, coronavirus pneumonia, and normal cases. Recent pneumonia detection methods were compared to the proposed architecture in terms of accuracy. The proposed approach provided robust and consistent pneumonia detection features with an accuracy rate of 99.81\% for radiography and 99.88\% for CT. This study's approach was confined to a single DL model, EfficientNetB7. 
Furthermore, the suggested architecture was assessed without using distributed or FL approaches, utilizing a traditional training and testing methodology for applying DL models.

To overcome the abovementioned limitations, this work offers a unique microservices and FTL-based approach that utilizes the learned knowledge, resulting in a robust and effective disease diagnosis model. The paper includes several contributions, such as proposing a microservices-based approach for disease diagnosis, improving privacy through FL, utilizing TL to improve model generalization,
and application in a real-world healthcare case study.
\section{Proposed Disease Detection Approach}
This study focuses on developing an intelligent method for disease detection through the utilization of distributed learning techniques. The proposed approach involves the implementation of a secure, flexible, scalable, and highly responsive system consisting of a collection of microservices. This architectural design enables efficient and effective disease detection utilizing distributed learning techniques while simultaneously ensuring the system's security, scalability, and responsiveness.

\begin{figure}[h]
\centering
\includegraphics[scale=0.6]{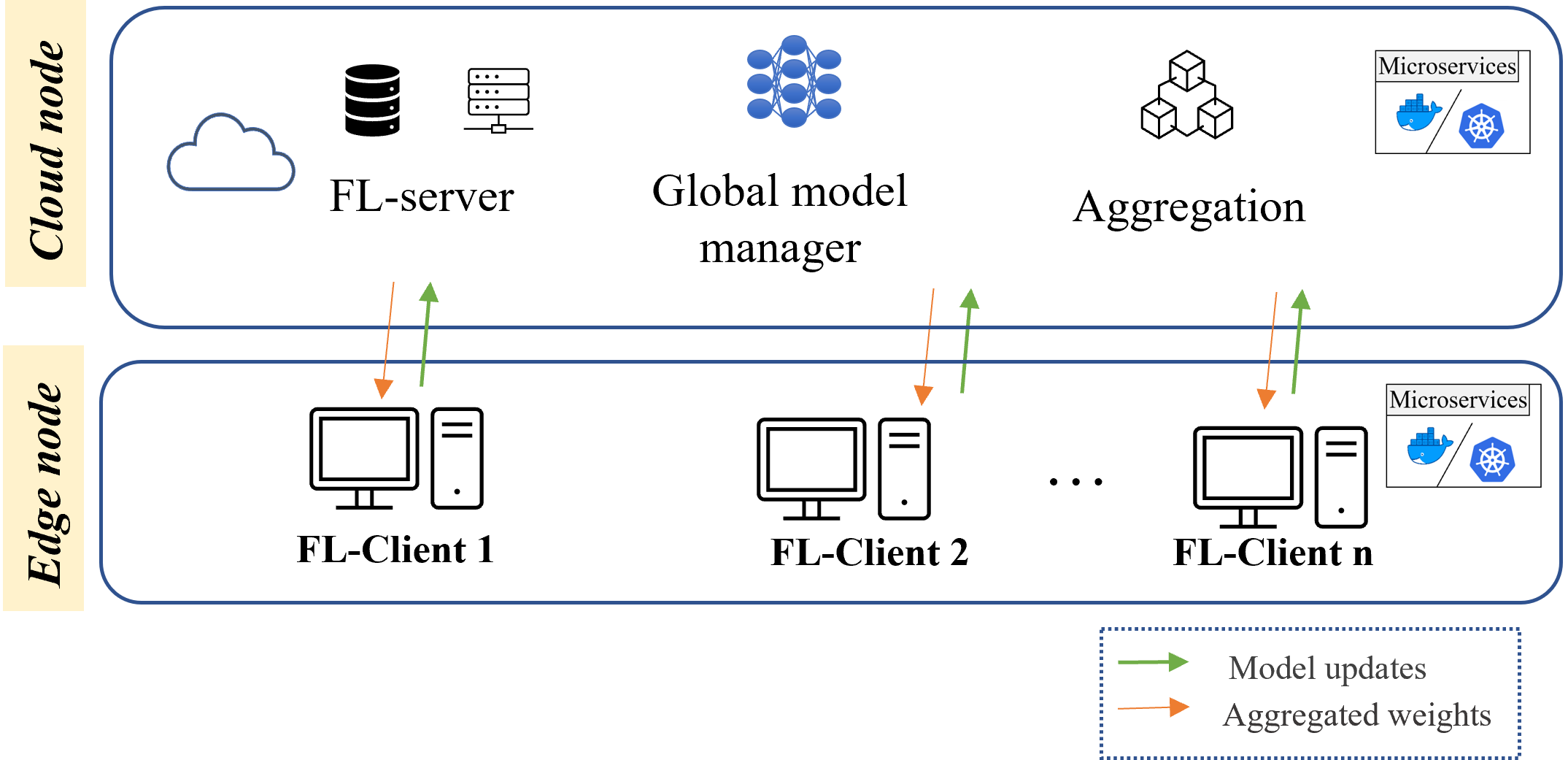}
\caption{The proposed distributed approach architecture for an efficient and secure disease detection process.}
\label{fig:pict1}
\end{figure}

The proposed approach involves cloud-edge collaborative architecture.
The architecture design of the proposed approach is presented in Figure \ref{fig:pict1} and showcases the integration of cloud and edge nodes.
The federated training process is seamlessly executed across these nodes, enabling efficient collaboration and leveraging the strengths of each component.

The first step in this process involves the collection of data from various sources, such as IoT devices, sensors, or other data-generating entities. This raw data is then transmitted to the edge computing resources, where preprocessing and local training occur.
Once the local training is completed, the model parameters are transmitted to the cloud FL server for aggregation. The cloud serves as a central hub for receiving and consolidating model updates from edge clients. Through the aggregation process, the cloud combines the knowledge learned from the various edge FL clients, creating a comprehensive and refined global model that benefits from the collective intelligence of the distributed network.



\begin{figure}[h]
\centering
\includegraphics[scale=0.6]{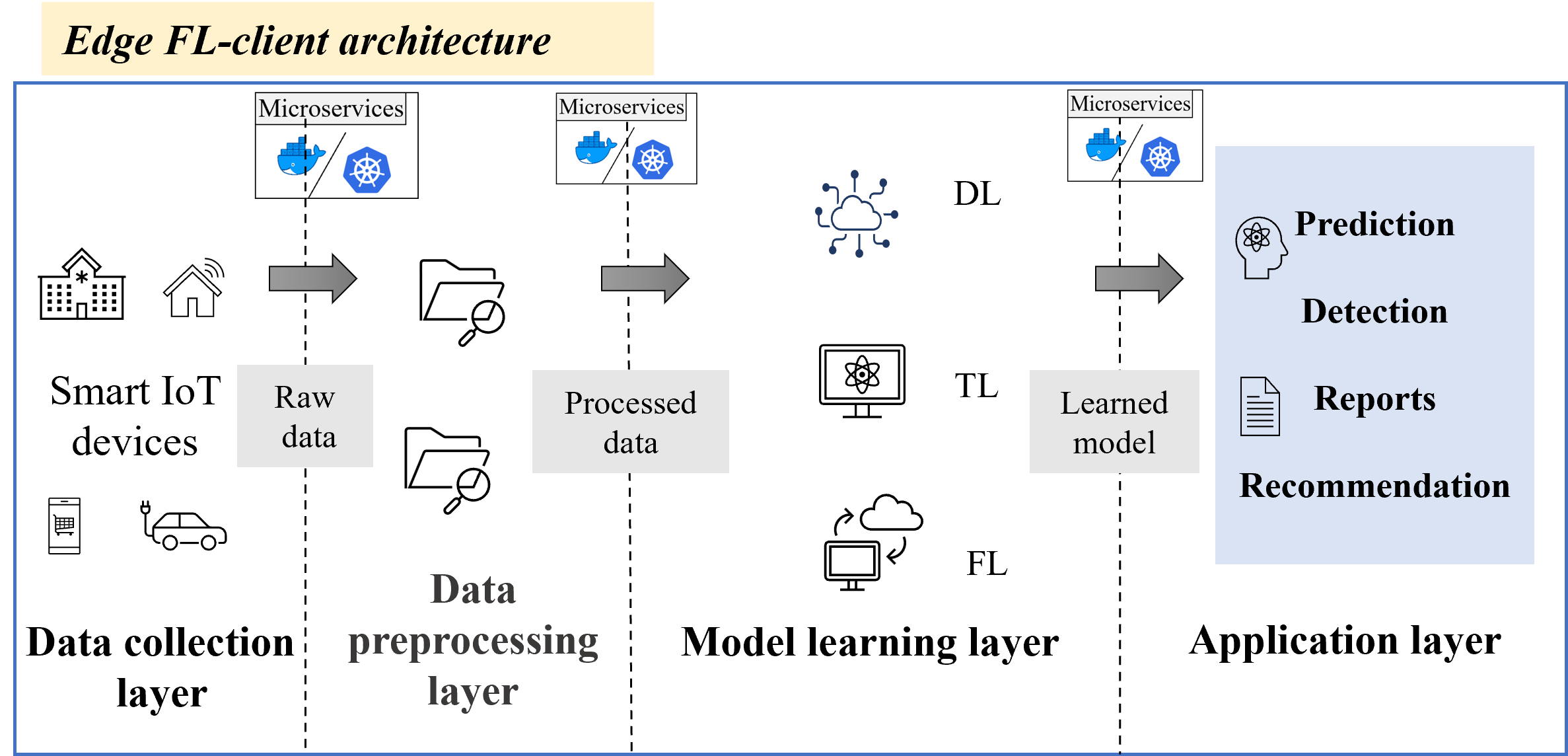}
\caption{The architecture of the FL client.}
\label{fig:pict22}
\end{figure}

The edge nodes in FL have a vital function by facilitating the training of DL models on data situated at the network's edge \cite{bao2022federated}.  This innovative approach utilizes an edge computer/server, eliminating the need to transmit data to a centralized server. By employing this process, data privacy in healthcare is significantly strengthened, and numerous benefits are achieved, including notable reductions in network latency and substantial savings in bandwidth consumption. In our proposed work, the edge node consists of a set of FL clients. Each FL client is formed by a set of layers 
and composed of service-oriented microservices that seamlessly collaborate to execute tasks, thereby empowering intelligent decision-making capabilities.

The FL client consists of four essential layers as illustrated in Figure \ref{fig:pict22}. The first layer is responsible for data collection, acquiring data from various sources located at the network's edge. Once the data is collected, the second layer, data preprocessing, performs necessary tasks to cleanse and prepare the data for subsequent stages. The third layer, model learning, establishes communication with the cloud FL server and initiates the training process based on predefined settings. This layer collaborates with cloud resources to effectively train and update the models. Finally, the application layer interprets the results generated by the trained models, providing insights and predictions.  




\subsection{Federated Leaning Procedure}

In the proposed distributed environment, multiple virtual instances at the network's edge are used to train models on their respective training data \cite{li2020federated}. The updated parameters are sent to a central cloud FL server for aggregation, creating a final global model incorporating insights from all individual instances. This global model is deployed on edge for use in analytics and applications.
The FL model development procedure follows a three-step method, which is outlined below.

\subsubsection{Model Selection and System Initialization}

The model manager selects the task,  like pneumonia detection, and generates suitable DL models. Multiple models can be chosen to be trained using the client's data, and TL techniques can be utilized to improve the performance of the models by leveraging previously acquired knowledge \cite{chen2020fedhealth}. The type of models to be used is determined based on the specific use case, and important parameters such as learning rates and communication rounds are selected to ensure effective and efficient training of the models. 
 
\subsubsection{Distributed Local Training and Updates}
After the model manager has set up the learning parameters and chosen the suitable models, the cloud FL server initiates the global models. It sends them to the FL clients to commence the learning process. Each client, labeled as k, trains its own local models using its own data, denoted as $D_{k}$, and calculates an update, denoted as $w_{k}$, by minimizing the loss function $F(w_{k})$:
\begin{equation}
    w_{k} = arg min F(w_{k} ), k \in K
\end{equation}
Afterward, each client k sends its computed update $w_{k}$ to the FL server for aggregation. The FL clients only exchange the model updates with the FL server throughout the training process, ensuring that data privacy is preserved.

\subsubsection{Models Aggregation and Download}
The server aggregates locally trained models from participating FL clients in each round, generating a new version of the global model. The number of communication rounds in FL refers to how often this aggregation technique is executed, involving the global model's collection, combination, and update. Each round typically includes multiple communication iterations where FL clients send their local models to the FL server, and the FL server sends back the updated global model. The number of communication rounds in the FL procedure is influenced by model convergence rate, dataset size, number of IoT devices, and communication bandwidth.

During each training cycle, the local updates provided by FL clients are merged using 
the federated averaging function 
, where the global model is generated by averaging the weights contributed \cite{mcmahan2017communication}.




\subsection{Hybrid Knowledge Transfer in Federated Learning}

Traditional Centralized Learning (CL) offers immense computational power but falls short of ensuring data privacy. To address this, FL has emerged as a privacy-preserving paradigm \cite{abdulrahman2020survey}. However, FL may face challenges in terms of computational efficiency compared to CL.
To further enhance model performance, we propose the integration of hybrid TL within FL. This approach combines the strengths of both TL and FL, enabling efficient knowledge transfer while safeguarding privacy.  
Hybrid TL leverages the power of pre-trained models, which have learned representations from large-scale centralized datasets, and combines them with FL to enhance model performance \cite{liu2020secure}. By employing TL techniques, models can benefit from existing knowledge while adapting to the specific data distributions present on distributed devices.

The hybrid TL approach in FL involves several steps. First, a base pre-trained model, which captures high-level features, is selected. 
Next, the base model is fine-tuned using FL on the local data of each client. Each FL client trains the model on its data while preserving privacy by only sharing model updates rather than raw data. The central FL server aggregates the model updates from all devices, allowing the global model to benefit from the collective knowledge of the devices.

In our proposed approach, we enhance the method of disease detection by incorporating different TL models. We employ popular CNN architectures, including VGG16, Xception, MobileNetV2, DenseNet201, and InceptionV3.


\subsection{Analytical Process in Microservices}

A collaborative ensemble of microservices is designed to work together, analyze data, and generate valuable insights for the data analytics process. These microservices are specifically designed and optimized for the main analytical tasks such as data preprocessing, model development, and interpretation. The microservices designed for each task are presented in the following:

\subsubsection{Data Preprocessing Stage}
The initial stage in data analysis is data preprocessing, which involves the following components:
\\
\textbf{- Data integration:} This component merges and arranges data from different sources to achieve a uniform format.
\\
\textbf{- Data scaling/transformation:} This component standardizes and modifies the collected data to a predefined range, aiming to improve data consistency. This step ensures that the data is reliable and suitable for subsequent analysis.

\subsubsection{Models Development Stage}
The model development stage is a crucial phase in the process of creating and refining models.  In our proposed approach, we adopt a collaborative approach where the model development takes place in conjunction with both the cloud server and the edge FL clients. The components deployed in the cloud are the following: 
\\
\textbf{- Model creator:} The model creator microservice generates and compiles the initial models before the training process. 
\\
\textbf{- Model uploader:} The model uploader microservice is responsible for securely and efficiently transferring the trained model parameters from the edge nodes to the central server. It ensures that the model parameters are transferred securely and reliably, maintaining the integrity and confidentiality of the model data.
\\
\textbf{- Model aggregator:} The model aggregator microservice in the cloud is a crucial component in FL. It receives the updated model parameters from the edge devices and combines them to generate a new global model.
\\
In the edge, these microservices are employed:
\\
\textbf{- Model training:} 
The model training microservice is responsible for conducting the training process according to the configuration parameters specified by the global model in the FL server. 
This microservice is deployed on the edge nodes.
\\
\textbf{- Model evaluator:} The model evaluator microservice assesses the performance of the model based on predefined metrics or criteria, such as accuracy, loss, or other relevant indicators.  

\subsubsection{Interpretation Stage}
In our approach, the last step in data analytics involves utilizing the learned models to make predictions or decisions using new data.
\\
\textbf{- Application microservice:} This microservice utilizes the trained models to make predictions or classifications when provided with new input data.


\section{Implementation and Experimental Analysis}
This section provides information about the dataset used for conducting the experiments. We also define the evaluation metrics employed to assess the performance of the proposed approach. Lastly, we present the results obtained from the experiments and conduct a detailed analysis of the findings in the last subsection.

\subsection{Dataset}
The dataset utilized for the experiments is named Pneumonia Chest X-ray Images and is publicly available \cite{PneuData}. It was collected by the Guangzhou Women and Children's Medical Center in China and can be found on Kaggle. The dataset contains a total of 5855 images, which are organized into three folders: train, validation, and test. The dataset is imbalanced, with a class distribution of approximately 30\% normal images and 70\% pneumonia images. 

\subsubsection{Experimental Setup}
The suggested solution was designed and validated using publicly accessible open-source components. It adopts a microservices-based architectural style along with FL and TL approaches, ensuring scalability and reliability. Docker containerization was used to create distributed images across three virtual machines, where each virtual machine hosted microservices of each data analytics stage using Docker instances \cite{docker}. Swarm orchestration managed the container cluster, providing high availability for applications \cite{kubernetes}. 
TensorFlow Federated (TFF) \cite{federated} and Keras \cite{keras}, open-source libraries, are used in the implementation of the models. These libraries offer powerful functionalities and tools for developing and training ML/DL models with the FL setting.

\subsection{Evaluation Metrics}
The evaluation metrics used in the proposed approach are as follows:
\\
\textbf{Accuracy:} It measures the model's overall performance across all categories and is calculated using Equation \ref{eq:4}. 
\\
\textbf{Precision:} It assesses the model's accuracy in classifying a sample as positive (pneumonia) or negative (normal) and is calculated using Equation \ref{eq:5}. 
\\
\textbf{Recall:} It measures the model's ability to identify positive samples and is calculated using Equation \ref{eq:6}. 
\\
\textbf{F1-score:} It combines accuracy and recall to provide a balanced performance measure and is calculated using Equation \ref{eq:8}. It considers both precision and recall.
\begin{equation} \label{eq:4}
    Accuracy=  \frac{TP+TN}{TP+TN+FP+FN}
\end{equation}
\begin{equation}\label{eq:5}
    Precision=  \frac{TP}{TP+FP}
\end{equation}   
 \begin{equation}\label{eq:6}
    Recall=  \frac{TP}{TP+FN}
\end{equation}        
\begin{equation}\label{eq:8}
    F1-score=  \frac{2*Precision*Recall}{Precision+Recall}
\end{equation}       
        
These metrics provide quantitative measures to evaluate the performance of the proposed approach in classifying normal and pneumonia chest X-ray images from the dataset. It is important to interpret these metrics collectively to comprehensively understand the model's performance and make informed decisions about its effectiveness for the studied task. 

\subsection{Experimental Results and Discussion}
 In this subsection, the results of the proposed approach for pneumonia detection are presented and analyzed. The outcomes of the implemented solution are discussed, and their implications are examined in detail.

 \subsubsection{Assessment of Approach: Analytics Findings for Pneumonia Detection}

 \begin{enumerate}
     \item \textbf{Data Preprocessing Stage}\\ 
      The dataset images were preprocessed by resizing them to a resolution of 224x224 and normalizing the pixel values to ensure consistency and compatibility across the dataset. In addition, we employed various techniques to augment the dataset, including re-scaling, horizontal flipping, random rotation, width and height shift, and adjusting zoom and brightness levels. These techniques were applied to enhance the diversity and variability of the dataset, leading to potentially improved model performance and generalization.
      \\In order to enable autonomous learning, the dataset was split into training and testing datasets, and these datasets were distributed among five separate clients.

   \item \textbf{Models Development}\\
The proposed approach was implemented using the TensorFlow Federated framework on 5 virtual IoT devices with a distributed configuration. Five pre-trained CNN architectures were utilized for TL to an efficient pneumonia detection, including VGG16, Xception, InceptionV3, ResNet50, and
DenseNet201. Models' aggregation was performed over 15 communication rounds, with each CNN trained for 20 epochs using local data. The Adam optimizer with a learning rate of 1e-4, the batch size of 32,  and the cross-entropy loss function were used for the parameter setting of DL models. 
    \item \textbf{Interpretation Stage}\\
After training the five CNNs, their performance was evaluated. The classification results of each model are shown in Table \ref{tab:comp}. The DenseNet201 model achieved the best performance, therefore it was selected to be deployed in the clients as the final global model for interpretation. The performance difference between the models is not significant, indicating relatively balanced performance. 
The performance outcomes of the generated models are presented in Table \ref{tab:results}. 

\begin{table}[h]
\centering
\caption{Performance results of the learned models using TL and FL. }
\label{tab:results}
\begin{tabular}{llllll}
\hline
\begin{tabular}[c]{@{}l@{}}Model//\\ Metric\end{tabular} & VGG16  & Xception  & ResNet50 & InceptionV3  & DenseNet201 \\ \hline
Accuracy    & 95.6 & 97.2 & 95.1 & 96.4 & 98.1 \\ \hline
Precision   & 95.4 & 97.4 & 94.6 & 96.8 & 97.9\\ \hline
Recall      & 96.2 & 97.9 & 95.3 & 96.5 & 98.3\\ \hline
F1-score    & 95.8 & 97.6 & 94.9 & 96.6 & 98.1\\ \hline
\end{tabular}
\end{table}

\end{enumerate}
 \subsubsection{Microservices Performance Assessment}
Moreover, the performance of the proposed approach was assessed by evaluating the execution time of microservices based on their functionality.
The end-to-end response time of the pneumonia detection function $f$ is defined as the sum of the time spent in data pre-processing $T_{Pre}$ and Interpretation $T_{Inter}$ 
stages, as shown in Equation \ref{eq:eqT1}. 
\begin{equation}
\label{eq:eqT1}
    T_{f} = T_{Pre} + T_{Inter} 
\end{equation}

\begin{figure}[h]
\centering
\includegraphics[scale=0.8]{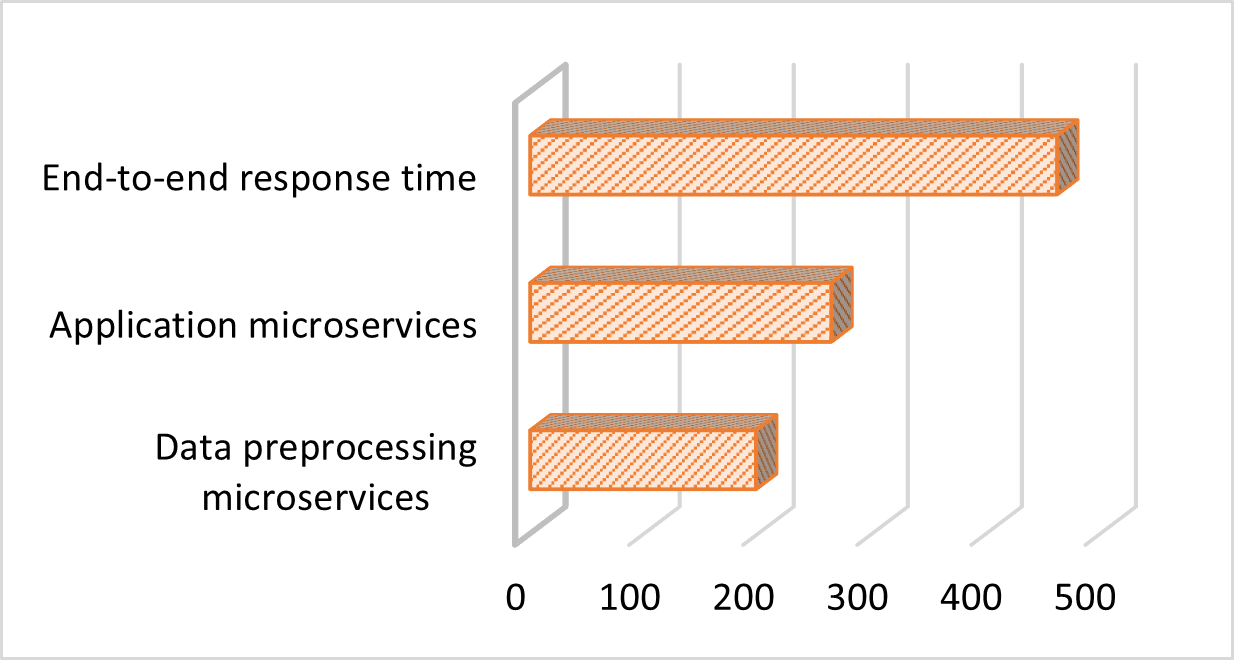}
\caption{The execution time of the data preprocessing and application microservices in addition to the end-to-end response time for the pneumonia detection process.}
\label{fig:pict2}
\end{figure}

Figure \ref{fig:pict2}  showcases the execution time of each microservice in the pneumonia detection process using the DenseNet201 model, encompassing data preparation and interpretation
. The data preprocessing microservice took 198 ms, while the interpretation microservice took 264 ms. As a result, the total time for the pneumonia detection process was calculated to be approximately 462 ms.
This rapid model responses enable healthcare professionals to make informed and data-driven decisions promptly. This can aid in selecting appropriate treatment plans and determining the urgency of interventions.

 \subsubsection{Comparison}
To validate the proposed approach, we compared our previously published study, which focused on the same use case scenario and utilized the same dataset but adopted a CL setting with monolithic architectures \cite{ben2022fusion}. The performance results, as depicted in Table \ref{tab:comp}, clearly demonstrate that our approach outperforms the previous study in terms of achieving the highest performance results. This validates the effectiveness of our approach that utilizes decentralized TL models in an FL architecture.

\begin{table}[h]
\centering
\caption{Comparison of the proposed approach results with our previously published study}
\label{tab:comp}
\begin{tabular}{lllllll}
\hline
Model & Type of learning & Architecture & Accuracy & Precision & Recall & F1-score  \\ \hline
\cite{ben2022fusion} & Centralized  & Monotholic & 97.5           &    97.5    &    98      &  97.8  \\ \hline
Proposed approach    & Distributed & Microservices &  98.1          &  97.9      &    98.3      &  98.1 \\ \hline
\end{tabular}
\end{table}

 \subsection{Discussion}
To ensure accurate pneumonia diagnosis, it is essential to identify all relevant features present in the chest X-ray images. 
The hybrid TL approach in FL offers several benefits. By leveraging pre-trained models, it harnesses the power of CL while preserving privacy through FL. The collective knowledge of the distributed devices enhances the model's performance. 

Privacy concerns arise when dealing with sensitive medical data. For this, FL was used to address these concerns by enabling the training of models on distributed data while keeping the data securely on their local devices. This allowed for collaborative model training across multiple institutions without sharing the raw data, preserving privacy and confidentiality.

In addition to privacy considerations, achieving high scalability and response time is crucial in healthcare. Microservices architecture was leveraged to address these requirements by breaking the data analytics function into smaller, specialized microservices that can be deployed and scaled independently. This allowed for efficient resource utilization and improved scalability. Moreover, microservices architecture enabled faster response times as each microservice can be optimized for performance and response time, leading to quicker results for end-users.

A robust and efficient pneumonia diagnosis solution was achieved by combining TL for high model performance, FL for privacy preservation, and microservices architecture for scalability and response time optimization. This approach ensures accurate predictions, protects patient privacy, and enables efficient and scalable data analytics in the healthcare environment.

\section{Conclusion}
This research aims to provide a novel approach for high-performing disease diagnosis models while ensuring privacy compliance. In detecting pneumonia using X-ray images, this solution uses TL for model development, FL for privacy preservation, and microservices architecture for scalability and response time optimization.

Our approach's experimental findings show that it can give accurate predictions, preserve patient privacy, and facilitate efficient and scalable data analytics in a healthcare setting. 

In the future, we aim to examine our approach's performance on various datasets to confirm its wide applicability and robustness in real-world case studies. In addition, we intend to develop and test fusion techniques, such as Dempster-Shafer-based fusion, weighted averaging, and attention-based strategies. These strategies can potentially improve the accuracy and robustness of the fusion process, and their effectiveness will be evaluated in the context of our approach.
In addition, we intend to explore and analyze the possible influence of various parameters on model performance, such as demographics, comorbidities, and disease severity, which might give useful insights into improving the suggested approach's overall performance.

\bibliography{references.bib}

\bibliographystyle{IEEEtran}

\end{document}